\documentclass[journal]{IEEEtran}
\ifCLASSINFOpdf
\else
\fi
\usepackage{url}

\usepackage{graphicx}
\DeclareGraphicsExtensions{.png}
\DeclareGraphicsExtensions{.pdf}
\usepackage{subcaption}
\usepackage{tikz}
\usetikzlibrary{shapes.geometric, arrows}
\usepackage{multirow}
\usepackage[flushleft]{threeparttable}

\tikzstyle{startstop} = [rectangle, rounded corners, minimum width=7.0cm, minimum height=0.8cm,text centered, text width=6.0cm, draw=black, fill=red!30]
\tikzstyle{io} = [trapezium, trapezium left angle=70, trapezium right angle=110, minimum width=5.5cm, minimum height=1.0cm, text centered, text width=6.0cm, draw=black, fill=blue!30, inner sep=10pt]
\tikzstyle{process} = [rectangle, minimum width=7.0cm, minimum height=0.8cm, text centered, text width=6.0cm, draw=black, fill=orange!30]
\tikzstyle{decision} = [diamond, minimum width=0.9cm, minimum height=0.9cm, text centered, text width=0.9cm, draw=black, fill=green!30]
\tikzstyle{arrow} = [thick,->,>=stealth]

\newcommand\copyrighttext{%
  \footnotesize \textcopyright 2020 IEEE. Personal use of this material is permitted. Permission from IEEE must be obtained for all other uses, in any current or future media, including reprinting/republishing this material for advertising or promotional purposes, creating new collective works, for resale or redistribution to servers or lists, or reuse of any copyrighted component of this work in other works.}
\newcommand\copyrightnotice{%
\begin{tikzpicture}[remember picture,overlay]
\node[anchor=south,yshift=10pt] at (current page.south) {\fbox{\parbox{\dimexpr\textwidth-\fboxsep-\fboxrule\relax}{\copyrighttext}}};
\end{tikzpicture}%
}

\begin{document}

\title{Deep Reinforcement Learning for Electric Transmission Voltage Control}

\author{Brandon~L.~Thayer,~\IEEEmembership{Member,~IEEE,}
        Thomas~J.~Overbye,~\IEEEmembership{Fellow,~IEEE,}% <-this % stops a space
\thanks{%This work has been submitted to the IEEE for possible publication. Copyright may be transferred without notice, after which this version may no longer be accessible.
This research was supported by funding from the Texas A\&M Engineering Experiment Station (TEES) Smart Grid Center (SGC). B. L. Thayer is with Pacific Northwest National Laboratory located in Richland, WA, USA. T. J. Overbye is with Texas A\&M University located in College Station, TX, USA.}% <-this % stops a space
}

% make the title area
\maketitle
\copyrightnotice

% As a general rule, do not put math, special symbols or citations
% in the abstract or keywords.
\begin{abstract}
Today, human operators primarily perform voltage control of the electric transmission system. As the complexity of the grid increases, so does its operation, suggesting additional automation could be beneficial. A subset of machine learning known as deep reinforcement learning (DRL) has recently shown promise in performing tasks typically performed by humans. This paper applies DRL to the transmission voltage control problem, presents open-source DRL environments for voltage control, proposes a novel modification to the ``deep Q network'' (DQN) algorithm, and performs experiments at scale with systems up to 500 buses. The promise of applying DRL to voltage control is demonstrated, though more research is needed to enable DRL-based techniques to consistently outperform conventional methods.

% This paper applies a subset of machine learning known as deep reinforcement learning (DRL) to the problem of steady state voltage control in the electric power transmission system. DRL environments that utilize PowerWorld Simulator were developed, and a state-of-the-art DRL algorithm that leverages ``deep Q networks’’ (DQN) was used. Recent DRL for voltage control experiments were reproduced, leading to algorithm and environment improvements. A novel change to the DQN algorithm where agents were not allowed to take the same action twice in any given episode led to significant improvements. It was found that using min-max scaled voltages in observations rather than per unit voltages led to improvements in successfully solving the voltage control problem. Experiments were carried out on 14, 200, and 500 bus systems to test scalability of DRL applied to voltage control. Results for the 200 and 500 bus systems were mixed, and instabilities were observed during training. For the 200 bus system, DRL agents were able to achieve success near to that of a heuristically-driven graph-based agent used for comparison. This paper demonstrates the promise of applying DRL to power system voltage control, but more research is needed to enable DRL-based techniques to consistently outperform conventional methods.
\end{abstract}

% Note that keywords are not normally used for peerreview papers.
\begin{IEEEkeywords}
Machine learning, artificial intelligence, automatic voltage control, power system simulation
\end{IEEEkeywords}

\IEEEpeerreviewmaketitle

\section{Introduction} \label{sec:intro}
\IEEEPARstart{T}{he} electric power grid is critical to the functioning of our modern society, and is undergoing a period of major change.
Large portions of the U.S. electric power system have undergone deregulation since the 1990's, distributing the responsibilities of electric power generation, transmission, and distribution to separate entities and opening up power markets \cite{pnnl_restructuring}.
Additionally, electricity generation from intermittent renewable resources such as wind and solar is rising while traditional generation sources such as coal and nuclear are declining \cite{eia_renewable_growth}.
Due to numerous factors including increasing extreme weather events, the reliability of the electric grid has been declining in recent years \cite{grid_reliability}. Further complicating these challenges are current and upcoming labor shortages in the electric power industry \cite{td_world_labor, incsys_operator}.

Fortunately, the digital computing revolution has been a boon to the electric power system.
Phasor measurement units (PMUs) have been increasing visibility into the electric power system, ``smart'' meter installations are on the rise, and energy management systems (EMS) are continually improving. Despite recent advances, many grid control actions are still taken by humans \cite{bpa_wide_area_control_1, ercot_operating_procedures, ercot_youtube, pjm_youtube}. As grid operation becomes increasingly complex, the automation of grid control is more important than ever.

To ensure secure grid operation, all buses in the transmission system are kept within a prescribed voltage band. Bus voltages are maintained by a variety of devices through a mixture of human and automatic control. The most prevalent voltage control devices are generators, switched shunts (\textit{e.g.}, capacitors), and on-load tap changing (OLTC) transformers \cite{nerc_reactive_power}. Generators typically follow a pre-determined voltage schedule created by the transmission or system operator, and voltage control is achieved by modulating reactive power output. Capacitors and voltage regulators may operate automatically based on local measurements, be directly controlled by human operators, and/or be controlled by a centralized optimization program \cite{bpa_wide_area_control_1, nerc_reactive_power, bpa_wide_area_control_2, italy_pt_1, pjm_voltage_control}. Recent advances in system-wide transmission voltage control leverage hierarchical control schemes and the use of ``pilot'' (bellwether) buses in pre-determined voltage control areas \cite{italy_pt_1, pjm_voltage_control, china_voltage_control}. While effective, these voltage control schemes make major modeling and control simplifications (\textit{e.g.}, linearization) due to their reliance on conventional optimization techniques. The large scale and complexity of the transmission system makes this optimization time consuming, which can render these techniques impractical when the need for rapid control decisions arises.

This paper presents the application of deep reinforcement learning (DRL) for automating transmission voltage control without requiring power system modeling simplifications. Open-source DRL environments were created, a novel modification to a popular DRL algorithm was made, and extensive experimentation was performed with 14, 200, and 500 bus test systems. The remainder of the paper is organized as follows: Section \ref{sec:background} provides background on DRL and its application to power system problems; Section \ref{sec:env-alg} describes the DRL environments created for this work, the DQN algorithm used, and a novel algorithm modification that led to improved performance; Section \ref{sec:graph-agent} describes random and graph-based algorithms for comparison with DRL; Section \ref{sec:14} presents extensive experimentation with the IEEE 14 bus system; Section \ref{sec:200-500} tests DRL scalability to larger power systems; and Section \ref{sec:conclusion} concludes the work.

\section{Background and Prior Work} \label{sec:background}
This section provides background on reinforcement learning and its application to power system control.

\subsection{Reinforcement Learning (RL) and ``Deep'' RL (DRL)}
Reinforcement learning (RL) is a form of learning in which an agent receives rewards or penalties for performing actions which affect its environment. Based on these rewards or penalties, an agent can potentially learn how to maximize its rewards according to the Bellman equation \cite{ai_book}. Q-learning is a model-free RL algorithm where agents attempt to learn the action-utility function (Q-function) that provides the expected utility (Q-value) of performing a particular action given the state of the environment. By learning the Q-function, the agent can then make decisions so as to optimize its cumulative future rewards. The learning of the Q-function often takes the form of a table where each entry represents the value of an action given the state of the environment \cite{ai_book}. This lookup table approach doesn't scale well to environments with large state and/or observation spaces, so recent advances use neural networks as an estimator of the Q-function. States are passed to the input layer of a neural network and an estimate of the expected utility of each possible action is emitted from the network's output layer. RL algorithms that leverage neural networks as Q-function estimators are collectively known as ``deep reinforcement learning'' (DRL) algorithms and have been proven successful in domains with large state and action spaces such as Atari video games and the board game Go \cite{google_dqn_original, google_nature, google_go}.

RL training and testing are broken up into ``episodes'' and ``time steps,'' wherein an episode consists of a discrete number of time steps. At each step, the agent receives an observation from the environment and a reward pertaining to their last action, and subsequently takes their next action. The action impacts the state of the environment, and the process continues. Episode initialization represents a ``reset'' of the environment, and episodes are typically terminated when the agent succeeds, fails, or exceeds a preset number of time steps.

\subsection{Reinforcement Learning for Power System Control}
The use of RL in the power system domain had been stymied by the large scale of states and control (actions) in the power system \cite{rl_grid_lit_review_old}. As DRL techniques have evolved and proven capable of operating in environments with larger state and action spaces, interest in RL's potential for power system control has increased. Reference \cite{rl_grid_lit_review} provides a literature survey of RL applied to electric power system control. Researchers have begun investigating RL for a variety of power system control problems including transient generator angle stability, congestion management, economic dispatch, and voltage control \cite{rl_grid_lit_review}. In \cite{rl_grid_lit_review}, it was suggested that the recent success of DRL may warrant a revisiting of previous grid control work where RL was applied before recent breakthroughs in the DRL domain.

One example of successfully revisiting grid control problems with new DRL algorithms is \cite{braking}. The work in \cite{braking} leveraged the open-source reinforcement learning environment framework known as Gym \cite{gym} as well as recently published open-source DRL algorithms \cite{baselines}, both created by OpenAI, in order to perform grid emergency control. OpenAI's Gym and DRL algorithms are discussed in more detail later on. In \cite{braking}, two grid control problems were investigated: dynamic generator braking and under-voltage load shedding. It was shown that reinforcement learning agents could be successfully trained both to apply a resistive generator brake in order to prevent the loss of generator synchronism and to shed the minimal amount of load required while maintaining a particular voltage recovery envelope. While the work presented in \cite{braking} is quite promising, the studies were performed using small test systems (10 and 14 buses).

The problem of power system reactive power/voltage control involves finding a solution to the power flow equations wherein all bus voltages stay within a prescribed band while reactive power reserves remain available for emergencies. Additional details pertaining the power flow equations can be found in \cite{wood}. In \cite{rl_voltage_old}, traditional, tabular Q-learning was applied to reactive power/voltage control. The system states were simplified by using a binary representation: if a component was within its operating limits (generator reactive power limits, transformer power flows, and bus voltages), the corresponding component of the state vector was 0. If a component was outside its operating limits, the corresponding state vector component was -1. The discrete action space consisted of transformer tap positions, shunt switch positions, and generator voltage set points. It was shown that Q-learning could be used to successfully determine control settings to reduce violations in 14 and 136 bus test cases.

\subsection{The ``GridMind'' Voltage Control Experiment} \label{sec:gridmind}
Similarly to \cite{rl_voltage_old}, \cite{gridmind} presented the application of reinforcement learning to reactive power/voltage control. However, newer DRL algorithms were used instead of tabular Q-learning. The reinforcement learning environment and agent in \cite{gridmind} were collectively referred to as ``GridMind.'' As the work in this paper significantly expands upon the findings of \cite{gridmind}, an extended description of \cite{gridmind} is presented in this section.

In \cite{gridmind}, per unit (p.u.) bus voltage magnitudes were used as observations for the deep reinforcement learning agent, and each individual action represented a set of voltage set points for all available generators (\textit{i.e.}, each action takes the form $\{v_{g_1}, v_{g_2} \dots v_{g_n}\}$ where $v_{g_1}$ is the voltage set point for generator $1$ and so on). Generator voltage set points were discretized into the set $\{0.95, 0.975, 1.0, 1.025, 1.5\}$. Note that this action definition leads to an action space with $n_{v}^{n_g}$ available actions, where $n_v$ is the number of discrete voltage set points and $n_g$ is the number of generators. The presented reward scheme gave a reward of $+100$ if all bus voltages were within $[0.95, 1.05]$ p.u., a penalty of $-50$ if any single bus voltage fell in either $[0.8, 0.95)$ or $(1.05, 1.25]$ p.u., or a penalty of $-100$ if any single bus voltage was $< 0.8$ or $> 1.25$ p.u. At the end of each episode, the agent received an additional reward (or penalty) equal to the mean of all rewards for the episode in question. 

The experiments performed in \cite{gridmind} used the IEEE 14 bus test system, which can be found at \cite{grid-cases}. The generators at buses 1 and 2 were considered to be available for active power dispatch, and the remaining three generators were available for voltage support only. Training episodes were created by randomly varying individual load levels between 80\% and 120\% of nominal without changing power factor. Tests were carried out with and without single line contingencies. The generators at buses 1 and 2 used participation factor control to adjust their active power set points. Episodes were terminated if all bus voltages fell within $[0.95, 1.05]$ p.u., the power flow failed to converge, or the agent reached a pre-determined per-episode action cap.

It's worth noting that in the absence of contingencies (all lines/transformers in service), loading levels of 80\%-120\% never result in low voltage conditions ($< 0.95$ p.u.) for the 14 bus system. In fact, at maximum loading (all loads at 120\% of nominal), the lowest voltage in the system is approximately 1.01 p.u. Conversely, the IEEE 14 bus base case has three generators set above the maximum acceptable voltage of 1.05 p.u. Thus, every episode begins with bus over-voltages.

The primary results presented in \cite{gridmind} were episode rewards as training progresses: at first, the agent did not earn very high rewards, but by the end of training, the agent could consistently earn high rewards and take very few actions.

The work presented in this paper builds on the methodology of \cite{gridmind}, and the experiments from \cite{gridmind} were reproduced as diligently as possible. Experiments were performed both with and without single line contingencies. All of the following discussion in this paragraph pertains to the reproduction of \cite{gridmind}, and code can be found at \cite{gym-powerworld, drl-powerworld}. Agents were trained for 500,000 simulation steps, and testing was performed on 5000 testing episodes that were not present in training. Without contingencies, the DRL agent could successfully bring all voltages into the acceptable band in 100\% of testing episodes. However, upon examining the actions taken by the agent in testing, it was found that only two of the available 3125 actions ($n_{v} = 5$, ${n_g} = 5$, $5^5 = 3125$) were utilized. In other words, the agent learned two sets of voltage set points that worked for every scenario. With single line contingencies included in each episode's initialization, the DRL agent was able to successfully bring all voltages in band in 83.8\% of testing episodes. Upon investigating the specific sequences of actions taken during testing, it was found that the agent can exhibit cyclic or repetitive behavior if it does not succeed in its first action. In other words, at times the agent took the same single action repeatedly until an episode terminated, and other times the agent would repeatedly cycle through identical sequences of actions until episode termination. Section \ref{sec:mod-alg} presents a novel algorithm modification that addresses this cyclic and repetitive behavior.

\section{DRL Algorithms and Environments for Voltage Control} \label{sec:env-alg}
This section contains details pertaining to the DRL algorithm used, enhanced environments created, and proposed algorithm modification for applying DRL to voltage control.

\subsection{Overview and DRL Algorithm Details} \label{sec:env-alg-overview}
Fig. \ref{fig:flowchart} presents a conceptual flowchart depicting DRL training for voltage control. Episode initialization (Fig. \ref{fig:flowchart}-\ref{flow:init}) can be adjusted to make an environment more or less ``challenging.'' Note that Fig. \ref{fig:flowchart} does not depict the inner workings of the DRL algorithm itself (Figs. \ref{fig:flowchart}-\ref{flow:compute-action}, \ref{fig:flowchart}-\ref{flow:take-action}), and the reader can find these details in \cite{google_dqn_original, baselines, drl-powerworld, stable-baselines, dueling_dqn, double_q_dqn, prioritized_replay_dqn}.

A software package was developed in the Python programming language for interfacing with PowerWorld Simulator, which was used to solve the power flow (Figs. \ref{fig:flowchart}-\ref{flow:start-ep}, \ref{fig:flowchart}-\ref{flow:implement-action}). This Python package, named EasySimAuto (ESA), significantly simplifies interfacing with Simulator's application programming interface (API) \cite{ESA}.

\newcounter{ac}
\renewcommand{\theac}{\alph{ac}}
\newcommand{\ac}[1]{(\refstepcounter{ac}\alph{ac})\label{#1}}

\begin{figure}[t]
\centering
\begin{tikzpicture}[node distance=1.3cm]
\tikzstyle{every node}=[font=\small]
\node (start) [startstop, label={[xshift=-0.7cm]right:\ac{flow:init}}] {Initialize all episodes (load levels, contingencies, generator set points, etc.).};
\node (start-ep) [process, below of=start, label={[xshift=-0.7cm]right:\ac{flow:start-ep}}] {Start new episode: Update loads, lines, generators, etc. Solve power flow.};
\draw [arrow] (start) -- (start-ep);
\node (initial-obs) [io, below of=start-ep, label={[xshift=-0.75cm]right:\ac{flow:initial-obs}}] {Send observations to agent (\textit{e.g.}, voltages).};
\draw [arrow] (start-ep) -- (initial-obs);
\node (compute-action) [process, below of=initial-obs, label={[xshift=-0.7cm]right:\ac{flow:compute-action}}] {Given observations, agent computes values of each potential action.};
\draw [arrow] (initial-obs) -- (compute-action);
\node (take-action) [io, below of=compute-action, label={[xshift=-0.75cm]right:\ac{flow:take-action}}] {Agent selects action with highest value.};
\draw [arrow] (compute-action) -- (take-action);
\node (implement-action) [process, below of=take-action, label={[xshift=-0.7cm]right:\ac{flow:implement-action}}] {Action implemented in the simulator. Solve power flow.};
\draw [arrow] (take-action) -- (implement-action);
\node (compute-reward) [process, below of=implement-action, label={[xshift=-0.7cm]right:\ac{flow:compute-reward}}] {Compute reward.};
\draw [arrow] (implement-action) -- (compute-reward);
\node (reward-obs) [io, below of=compute-reward, label={[xshift=-0.75cm]right:\ac{flow:reward-obs}}] {Send new observation and reward to agent.};
\draw [arrow] (compute-reward) -- (reward-obs);
\node (ep-over) [decision, below of=reward-obs, yshift=-0.3cm, xshift=-2.1cm] {Episode over?};
\draw [arrow] (reward-obs) -- (ep-over);
\draw [arrow] (ep-over.west) -| node[anchor=west, yshift=+0.2cm, xshift=+0.2cm] {yes} ++(-1, 0) |- (start-ep);
\node (train-over) [decision, below of=reward-obs, yshift=-0.3cm, xshift=+2.1cm] {Training over?};
\draw [arrow] (ep-over.east) -- node[anchor=east, yshift=+0.2cm, xshift=-0.2cm] {no} (train-over);
\node (stop) [startstop, below of=train-over, xshift=-2.0cm, yshift=-0.3cm] {Stop.};
\draw [arrow] (train-over) -- node[anchor=south, xshift=-0.2cm] {yes} (stop);
\draw [arrow] (train-over.east) -| node[anchor=east, yshift=+0.2cm, xshift=-0.2cm] {no} ++(+1, 0) |- (compute-action);
\end{tikzpicture}
\caption{Conceptual Flowchart: DRL for Voltage Control} \label{fig:flowchart}
\end{figure}
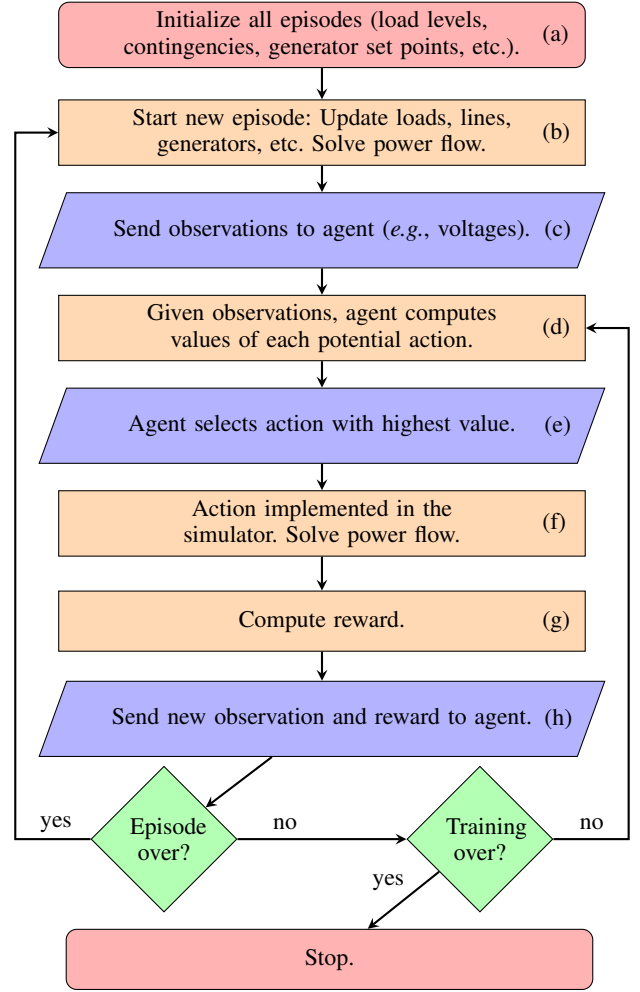

A Python package called ``Gym,'' created by OpenAI, is an open-source toolkit for developing and comparing RL algorithms, and is commonly used in RL research as a way to standardize RL environment development \cite{gym}. For the work presented here, a set of several environments were constructed, and the software repository can be found at \cite{gym-powerworld}. The environments use ESA to send commands to and retrieve data from the power flow simulator (Figs. \ref{fig:flowchart}-\ref{flow:start-ep}, \ref{fig:flowchart}-\ref{flow:initial-obs}, \ref{fig:flowchart}-\ref{flow:implement-action}, \ref{fig:flowchart}-\ref{flow:reward-obs}). The environment itself is additionally responsible for initialization (Fig. \ref{fig:flowchart}-\ref{flow:init}) and reward computation (\ref{fig:flowchart}-\ref{flow:compute-reward}).

A collection of high-quality DRL algorithm implementations is provided by \cite{baselines} (also from OpenAI). An improved and documented version of the algorithms in \cite{baselines} is provided by \cite{stable-baselines}, and was leveraged in this work. Specifically, an advanced DRL algorithm with so-called ``deep Q networks'' (DQN) originally presented in \cite{google_dqn_original} was used. Several modifications/improvements to the algorithm have been published over the years including ``dueling DQN'' \cite{dueling_dqn}, ``double-Q learning'' \cite{double_q_dqn}, and prioritized experience replay \cite{prioritized_replay_dqn}. The work here used all the aforementioned DQN algorithm improvements. DQN algorithms require discrete action spaces, so continuous control elements such as generator voltage set points must be discretized. The DQN algorithm with all improvements has many different hyper parameters which can be tuned. In this work, the majority of hyper parameters were left at their default values, and are explicitly defined in \cite{drl-powerworld}.

\subsection{DRL Environment Details}
As discussed in Section \ref{sec:gridmind}, the GridMind experiment varied loads between 80\% and 120\% of nominal, varied generator active power output linearly from the base case settings, and used constant generator voltage set points for all episodes. By contrast, this section presents details on a significantly more challenging environment which can be used to test the limits of DRL applied to transmission voltage control. The environment described here was used along with the DQN algorithm and all improvements (see Section \ref{sec:env-alg-overview}) to generate the results presented in Sections \ref{sec:14} and \ref{sec:200-500}.

\subsubsection{Episode Initialization Overview} \label{sec:general_ep_init}
The design of how episodes (scenarios) are created can have a major impact both on a reinforcement learning agent's ability to learn and on the ultimate usefulness of a trained agent. For instance, training under a narrow range of load and generator conditions may enable an agent to learn the best control actions quickly, but the agent's decision making may not generalize well to other conditions not encountered in training.

\subsubsection{Episode Initialization - System Loading} \label{sec:load_ep_init}
For each scenario, total system active power loading is drawn from the uniform distribution between 60\% and 140\% of the base case's total active power loading. Next, a value from the uniform distribution on the interval $[0,1)$ is independently drawn for each load in the system. These values are subsequently summed and then linearly scaled such that after scaling the values sum to one. After this scaling, the new values for each load represent the load's fraction of total system active power loading ($P$). In this way, each load can theoretically take on a value between 0\% and 100\% of the given episode's total system loading (though the extremes are incredibly unlikely to occur). Next, each load has a power factor ($p\! f$) drawn from the uniform distribution on the interval $[0.8, 1.0)$, and reactive power levels ($Q$) are computed via the relationship $Q = P \cdot \tan(\arccos(p\! f))$. Finally, each load has a 10\% chance for its power factor to be changed from lagging to leading (flipping the sign of $Q$).

\subsubsection{Episode Initialization - Generator Active Power} \label{sec:gen_active_power_init}
After computing system loading for each scenario, the generator commitment can be determined and active power levels can be dispatched to meet demand. Since generators have active power output minimum and maximum allowable values, the procedure for determining generation levels differs somewhat from the procedure for determining individual loads. The following description is functionally equivalent to what is done in the environment code \cite{gym-powerworld}, but is explained in a simplified manner. The actual code is completely vectorized.

For each scenario, a random ordering of all generators in the case is drawn. Then, the generators are looped over in the given random order, and an active power output is drawn from the uniform distribution between the particular generator's $P_{min}$ and $P_{max}$. This process continues until the total active power output of the generators meets or exceeds the total loading for the given scenario, plus assumed losses of 3\%. In this way, different generators may be active for each scenario, effectively building in generator contingencies and creating different unit commitments for each scenario. Assuming some losses are present ensures the slack generator must not cover all active power losses, which can be significant depending on the number of lines and their resistance.

\subsubsection{Episode Initialization - Generator Voltage Set Points} \label{sec:gen_voltage_init}
Random voltage regulation set points are drawn uniformly for each episode and each generator from the set $\{0.95, 0.975, 1.00, 1.025, 1.05\}$ p.u.

\subsubsection{Episode Initialization - Lines and Shunts}
For cases which contain shunts, initial shunt states (open/closed) are simply randomly drawn from the uniform distribution for all shunts and all episodes. A single branch (line or transformer) is randomly removed from service for each episode.

\subsubsection{Observation Design} \label{sec:observation-design}
Engineering the observations given to the DRL agent is critical to successful learning, and requires significant experimentation. It's important to ensure the appropriate amount of information is given to the agent - too little information and the agent may fail to learn due to a lack of observability, while too much information can significantly scale up the size of the neural network required for DRL and cause a failure to learn due to increased difficulty in finding relationships between features.

There are many different options available for configuring the environments described in this work \cite{gym-powerworld}. The following observation combinations are used in Sections \ref{sec:14} and \ref{sec:200-500}:
\begin{itemize}
    \item Bus voltage magnitudes only
    \item Bus voltage magnitudes and gen. states (on/off)
    \item Bus voltage magnitudes and branch states (open/closed)
    \item Bus voltage magnitudes, gen. states, and branch states
    \item Bus voltage magnitudes, gen. states, branch states, and shunt states (open/closed)
\end{itemize}

All of the observation combinations above can also be used with transformed bus voltage magnitudes. In these cases, voltages are transformed via ``min-max'' scaling on the interval $[0, 1]$. The environments consider a power flow to be ``failed'' if any single bus voltage is $ < 0.7$ p.u. or $ > 1.2$ p.u., even if the simulator is able to solve the power flow. In this way, the absolute lower and upper voltage bounds are known ahead of time for all scenarios/episodes, enabling min-max scaling.

\subsubsection{Action Space}
As mentioned in Section \ref{sec:gridmind}, the action space presented in \cite{gridmind} scales as $n_{v}^{n_g}$. If voltage set points are discretized into five settings and five generators are present (as in the IEEE 14 bus system), the action space has dimension 3125. This clearly does not scale well to larger systems. By contrast, the environments in this work map each voltage setting for each generator to a single action, resulting in an action space that scales as $n_{v} \cdot n_{g}$. For systems with switchable shunts (namely capacitors), a single action is available per shunt that toggles the shunt state (open/closed). In all cases, a ``no-op'' action is included which does not lead to any change in the system.

\subsubsection{Reward Design Overview}
In some of the recent RL for power systems literature, agents were primarily rewarded based on the post-action state of the system, rather than the \textit{change} that the given action \textit{induced} in the system \cite{braking, gridmind}. If the objective is to bring all bus voltages within the range $[0.95, 1.05]$ per unit, an action that moves a bus voltage from 0.93 to 0.945 p.u. should be rewarded despite the voltage not moving into the acceptable band, while an action that moves a bus voltage from 0.96 to 0.94 p.u. should be penalized. Two movement-based control schemes are proposed in this work. Readers seeking details beyond what is presented in the following two sections can consult \cite{gym-powerworld}.
% For brevity, this paper does not provide all of the mathematical equations pertaining to these reward schemes. However, additional details can be found in \cite{gym-powerworld}.

\subsubsection{Reward Scheme 1}
This reward scheme provides rewards for moving voltages in the right direction (toward the acceptable band), while providing penalties for voltages that move in the wrong direction (away from the acceptable band). The movement rewards and penalties are scaled by the movement magnitude so that a larger voltage movement obtains a larger reward or penalty. Additionally, a penalty is given for taking an action in order to help incentivize the agent to minimize the number of actions taken.

\subsubsection{Reward Scheme 2}
This scheme keeps all rewards within the range $[-1, +1]$, inspired by the reward ``clipping'' done in \cite{google_dqn_original, google_nature}. In \cite{google_dqn_original, google_nature} it was noted that keeping rewards in a fixed band ``limits the scale of the error derivatives'' and thus makes the reward function useful across multiple games. In the case of the work presented in this paper, the same notion applies, but instead helps the reward scheme generalize across power system cases. The clipped scheme presented here has the following discrete reward possibilities: $\{-1.00, -0.75, -0.50, -0.25, -0.10, 0.00, +0.25, +0.50,\\+0.75, +1.00 \}$. A reward of $-1$ is given if the power flow diverges, and a reward of $+1$ is given if all bus voltages are in band. Other rewards (penalties) are given based on the net number of buses that move in the right (wrong) direction.

\subsection{DQN Algorithm Modification} \label{sec:mod-alg}
As mentioned in Section \ref{sec:gridmind}, the unmodified DQN algorithm can select the same action multiple times in a given episode.
% While this may be sensible for video games, it isn't in the context of power system voltage control.
One novel contribution of this work is the modification of the DQN algorithm such that each action is only permitted to be taken once per episode. If, during the course of training or testing, the agent determines an action which has already been taken in the current episode to have the highest value, the action with the second highest value is chosen instead. This process continues until an action which has not been taken yet is selected. 

The aforementioned algorithm modification is useful in training as the agent is forced to perform additional exploration. Additionally, this modification makes sense in the context of power system voltage control: repeatedly commanding a capacitor closed or sending a generator voltage set point will not improve system voltage conditions. The experiments presented in Section \ref{sec:14} show that this modification leads to significantly higher success rates in the environments described in this section. The code for the modification can be found at \cite{drl-powerworld}.

\section{Random and Graph-Based Agents \\For Comparison} \label{sec:graph-agent}
In order to provide a basis for comparison with DRL results, both a random agent and a graph-based agent were developed. Both agents interact with the environments described in Section \ref{sec:env-alg} and attempt to solve the voltage control problem. All testing is performed against the exact same testing episodes that the DRL agents were tested against (Sections \ref{sec:14} and \ref{sec:200-500}). As per usual, each episode proceeds either until all voltage issues have been fixed, the power flow diverges, any bus voltage goes below 0.7 p.u. or above 1.2 p.u., or the agent hits the per-episode action cap.

The random agent behaves exactly as one might expect: during each time step it randomly chooses and takes an action from the environment's action space. The random agent is run both with and without the unique-actions-per-episode requirement. For RL generally, it is considered best practice to compare results with a random agent as a baseline to ensure that the RL algorithm is functioning as intended.

The graph-based agent has been created to provide a reasonable benchmark for comparison with the DRL agents. In contrast to the DRL agent, the graph-based agent uses a model of the power system and is heuristically driven, leveraging the notion that voltage issues are typically ``local'' and are often remedied by dispatching reactive power resources near to the buses with voltage issues. In short, the agent constructs a weighted, undirected graph of the power system network and changes the voltage set point at the generator which is ``nearest'' to the bus with the minimum/maximum voltage violation. Graph weights are defined as branch reactances, and ``nearest'' is defined as the shortest path length considering reactances as distances. For cases that contain capacitors, capacitors at neighboring buses to the bus with the largest violation are actuated appropriately (opened for high voltage violations, closed for low voltage violations) before generator set points are considered. The code for the graph-based agent can be found in \cite{drl-powerworld}.

\section{Experiments with IEEE 14 Bus System} \label{sec:14}
This section presents results for DRL agents attempting to control voltage in the IEEE 14 bus system. All neural networks are fully connected and contain two hidden layers with 64 neurons each. The environment and DQN algorithm with all improvements are as described in Section \ref{sec:env-alg}. Tests were performed both with and without the DQN algorithm modification discussed in Section \ref{sec:mod-alg}. In contrast to what was done in \cite{gridmind} (Section \ref{sec:gridmind}), all five generators are considered available for active power dispatch (Section \ref{sec:gen_active_power_init}). All episodes contain a single line contingency: the lines considered for contingencies were (from bus - to bus) 1-5, 2-3, 4-5, and 7-9. All training sessions were allowed to proceed for 500,000 simulation steps (total actions taken by agent), and all episodes were capped at 10 actions (twice the number of generators). Note the 14 bus system does not contain any capacitors. Testing is performed on 5000 testing episodes not seen in training.

Table \ref{table:all-14} depicts results grouped by observations provided to the agents. Note that the abbreviated column headers are explicitly defined at the bottom of Table \ref{table:all-14}. All success percentages and rewards represent the average across three independent training/testing runs with different random number generator seeds. The column labeled ``PSOOBV'' presents the success rate for testing episodes which began with out-of-band (OOB) voltages (outside $[0.95, 1.05]$). For comparison, the graph-based and random agents obtained OOB success rates of 40.8\% and 16.5\%, respectively. All OOB success rates lower than the random agent are \textit{italicized} for emphasis. Across all four observation groupings in Table \ref{table:all-14}, the combination of the unique actions per episode (UAE) algorithm modification, min-max scaled voltages (MMV), and reward scheme (RS) 1 generally led to the best results (values in these rows are \textbf{bold}). Success rates for this combination were very similar across observation groupings. While this aforementioned combination was best, the graph-based agent still performed better with its OOB success rate of 40.8\%. 

\begin{table}[h]
\centering
\begin{threeparttable}
\caption{Results for 14 bus experiments}
\label{table:all-14}
\begin{tabular}{|l|l|l|l|l|l|}
\hline
\textbf{Observations} & \textbf{UAE} & \textbf{MMV} & \textbf{RS} & \textbf{PS} & \textbf{PSOOBV} \\ \hline
\multirow{5}{*}{Voltage Only}

    & No  & No  & 1 & 13.4 & \textit{6.0} \\ \cline{2-6} 

    & Yes & No  & 1 & 38.5 & 31.8 \\ \cline{2-6} 

    & No  & Yes & 1 & 19.4 & \textit{11.6} \\ \cline{2-6} 

    & \textbf{Yes} & \textbf{Yes} & \textbf{1} & \textbf{42.8} & \textbf{36.2} \\ \cline{2-6} 

    & Yes & Yes & 2 & 14.8 & \textit{9.6} \\ \hline \hline

\multirow{5}{*}{\begin{tabular}[c]{@{}l@{}}Voltage and\\ Gen. State\end{tabular}}                    & No  & No  & 1 & 19.4 & \textit{12.8} \\ \cline{2-6} 
                                                                                                     & Yes & No  & 1 & 31.7 & 24.9 \\ \cline{2-6} 
                                                                                                     & No  & Yes & 1 & 21.0 & \textit{14.5} \\ \cline{2-6} 
                                                                                                     & \textbf{Yes} & \textbf{Yes} & \textbf{1} & \textbf{41.6} & \textbf{35.7} \\ \cline{2-6} 
                                                                                                     & Yes & Yes & 2 & 16.4 & \textit{11.2} \\ \hline \hline
\multirow{5}{*}{\begin{tabular}[c]{@{}l@{}}Voltage and\\ Branch State\end{tabular}}                  & No  & No  & 1 & 16.9 & \textit{8.9} \\ \cline{2-6} 
                                                                                                     & Yes & No  & 1 & 37.4 & 30.5 \\ \cline{2-6} 
                                                                                                     & No  & Yes & 1 & 17.3 & \textit{9.7} \\ \cline{2-6} 
                                                                                                     & \textbf{Yes} & \textbf{Yes} & \textbf{1} & \textbf{41.3} & \textbf{34.6} \\ \cline{2-6} 
                                                                                                     & Yes & Yes & 2 & 14.6 & \textit{8.7} \\ \hline \hline
\multirow{5}{*}{\begin{tabular}[c]{@{}l@{}}Voltage,\\ Gen. State,\\ and Branch\\ State\end{tabular}} & No  & No  & 1 & 19.5 & \textit{12.8} \\ \cline{2-6} 
                                                                                                     & Yes & No  & 1 & 40.8 & 35.4 \\ \cline{2-6} 
                                                                                                     & No  & Yes & 1 & 20.1 & \textit{13.9} \\ \cline{2-6} 
                                                                                                     & \textbf{Yes} & \textbf{Yes} & \textbf{1} & \textbf{40.9}  & \textbf{35.1} \\ \cline{2-6} 
                                                                                                     & Yes & Yes & 2 & 16.0 &\textit{11.1} \\ \hline
\end{tabular}
\begin{tablenotes}
\small
\item \textbf{UAE}: Unique Actions per Episode, \textbf{MMV}: Min-Max Voltages, \textbf{RS}: Reward Scheme, \textbf{PS}: Percent Success, \textbf{PSOOBV}: Percent Success, Out-of-Band Voltages
\end{tablenotes}
\end{threeparttable}
\end{table}

It can also be seen in Table \ref{table:all-14} that the use min-max scaled voltages generally led to better results than p.u. voltages (``No'' in the MMV column). This is due to the fact that min-max scaling magnifies differences in voltages and makes it easier for the agent to learn. Table \ref{table:all-14} also shows that reward scheme 2 consistently led to worse results than the random agent.

Due to the simplicity of the 14 bus system, it is hypothesized that the graph-based agent's success rate of 40.8\% approaches the actual percentage of episodes in which it was physically possible to bring all voltages in band (limited by generator reactive power limits). In this light, it can be seen that in several cases the DRL agents performed relatively well.

\section{Experiments with 200 and 500 Bus Systems} \label{sec:200-500}
In this section, the findings from Section \ref{sec:14} are exploited to test the scalability of DRL for voltage control to significantly larger and more realistic systems. Synthetic grid test cases with 200 and 500 buses presented in \cite{synthetic_grids} and available at \cite{grid-cases} are leveraged. These synthetic grids have been constructed to be representative of the real electric grid, but are intentionally different in order to avoid data sensitivity issues.

The 200 bus system contains 49 generators, four switched shunts, and 180 transmission lines, while the 500 bus system contains 90 generators, 17 shunts, and 468 lines.

Training and testing was performed as in Section \ref{sec:14}, but training was allowed to proceed for 2,000,000 and 3,000,000 time steps for the 200 and 500 bus systems, respectively. Experiments used fully connected neural networks with two hidden layers. Hidden layers contained 1024 neurons each for 200 bus experiments, and 2054 neurons for 500 bus experiments. Since out-of-band (OOB) success rates for the various observation groups listed in Table \ref{table:all-14} were similar and these systems are more complex than the 14 bus system, agents were given min-max scaled voltages, generator states (on/off), and shunt states (open/closed) as observations. Each system was tested both with and without single line contingencies. If contingencies were included, the agent is additionally given all line states (open/closed) as observations.

\begin{figure}[h!]
  \begin{subfigure}[b]{\linewidth}
    \centering
    \includegraphics[width=\linewidth]{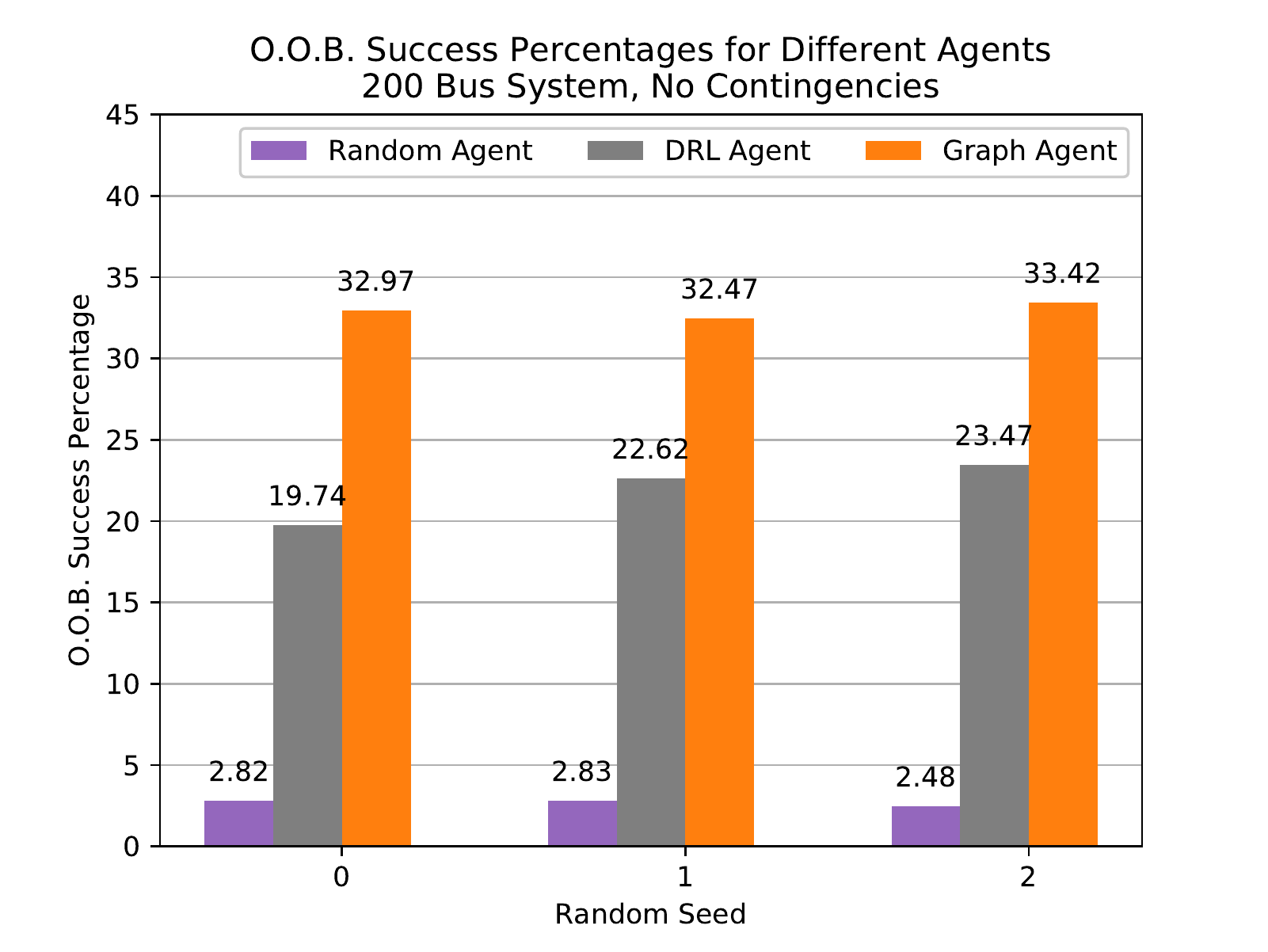} 
    \caption{No contingencies} 
    \label{fig:200-no-con} 
    % \vspace{1ex}
  \end{subfigure}
  \begin{subfigure}[b]{\linewidth}
    \centering
    \includegraphics[width=\linewidth]{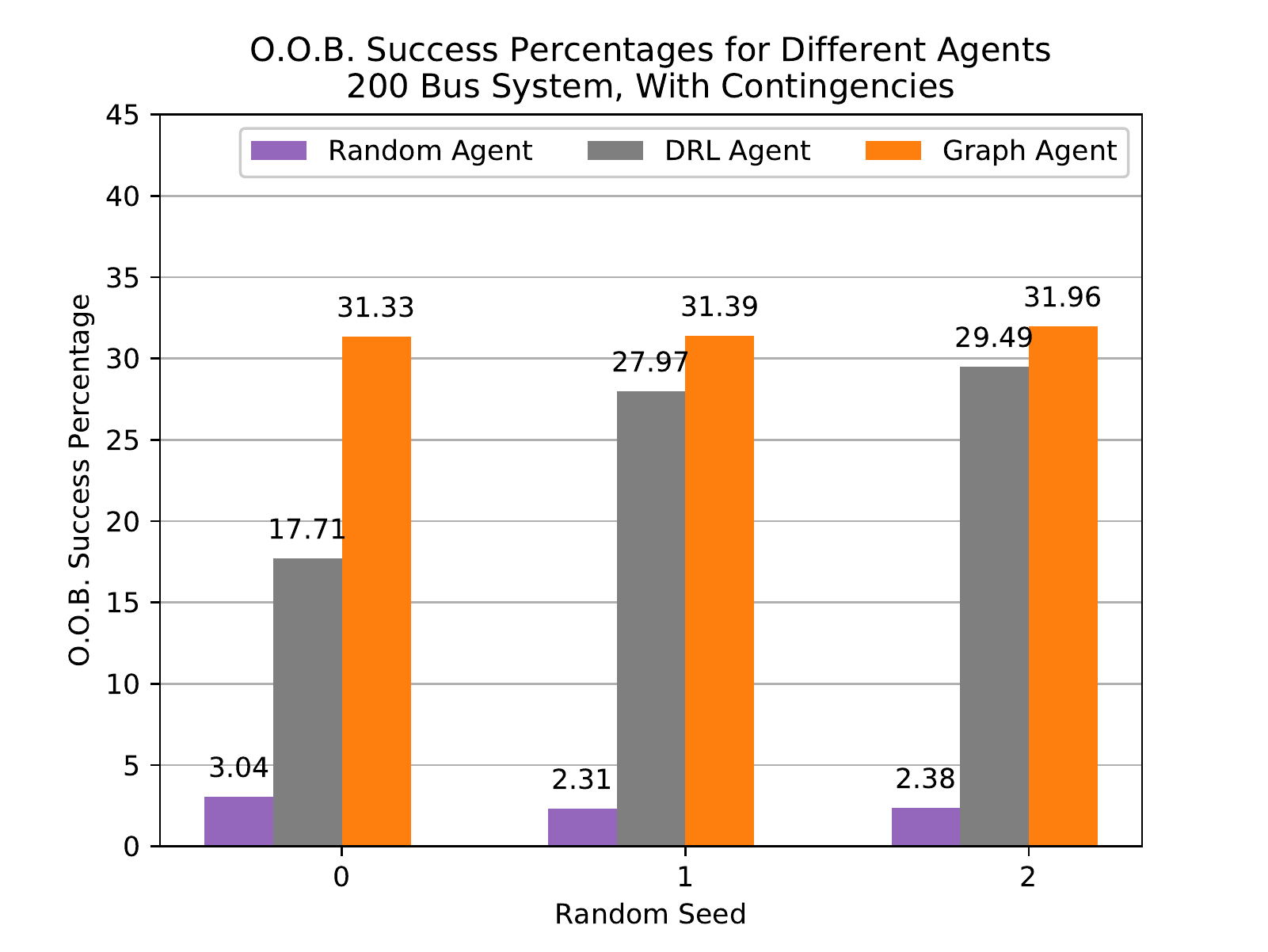} 
    \caption{With contingencies} 
    \label{fig:200-with-con} 
    % \vspace{1ex}
  \end{subfigure} 
  \caption{OOB success rates for random, DRL, and graph-based agent across different random seeds, 200 bus system, with and without contingencies}
  \label{fig:200-results} 
\end{figure}

Fig. \ref{fig:200-results} presents OOB success rates for random, DRL, and graph-agents across three independent runs with different random seeds for the 200 bus system. Fig. \ref{fig:200-no-con} shows success rates when no contingencies are applied in the system, and Fig. \ref{fig:200-with-con} shows success rates when single line contingencies are applied for each episode. Similarly to the results in Section \ref{sec:14}, the graph-based agent consistently outperforms the DRL agents, and the DRL agents consistently outperform the random agents. However, it can be seen in Fig. \ref{fig:200-with-con} that the DRL agents approached the performance of the graph-based agent for two out of three experimental runs. One key takeaway from Fig. \ref{fig:200-results} is that the DRL agent's performance on the whole was actually better when single line contingencies were applied. It is hypothesized that this is a result of an oversized neural network in the case of no contingencies: when contingencies are included the observation space grows from 253 measurements to 433.

\begin{figure}[h!]
  \begin{subfigure}[b]{\linewidth}
    \centering
    \includegraphics[width=\linewidth]{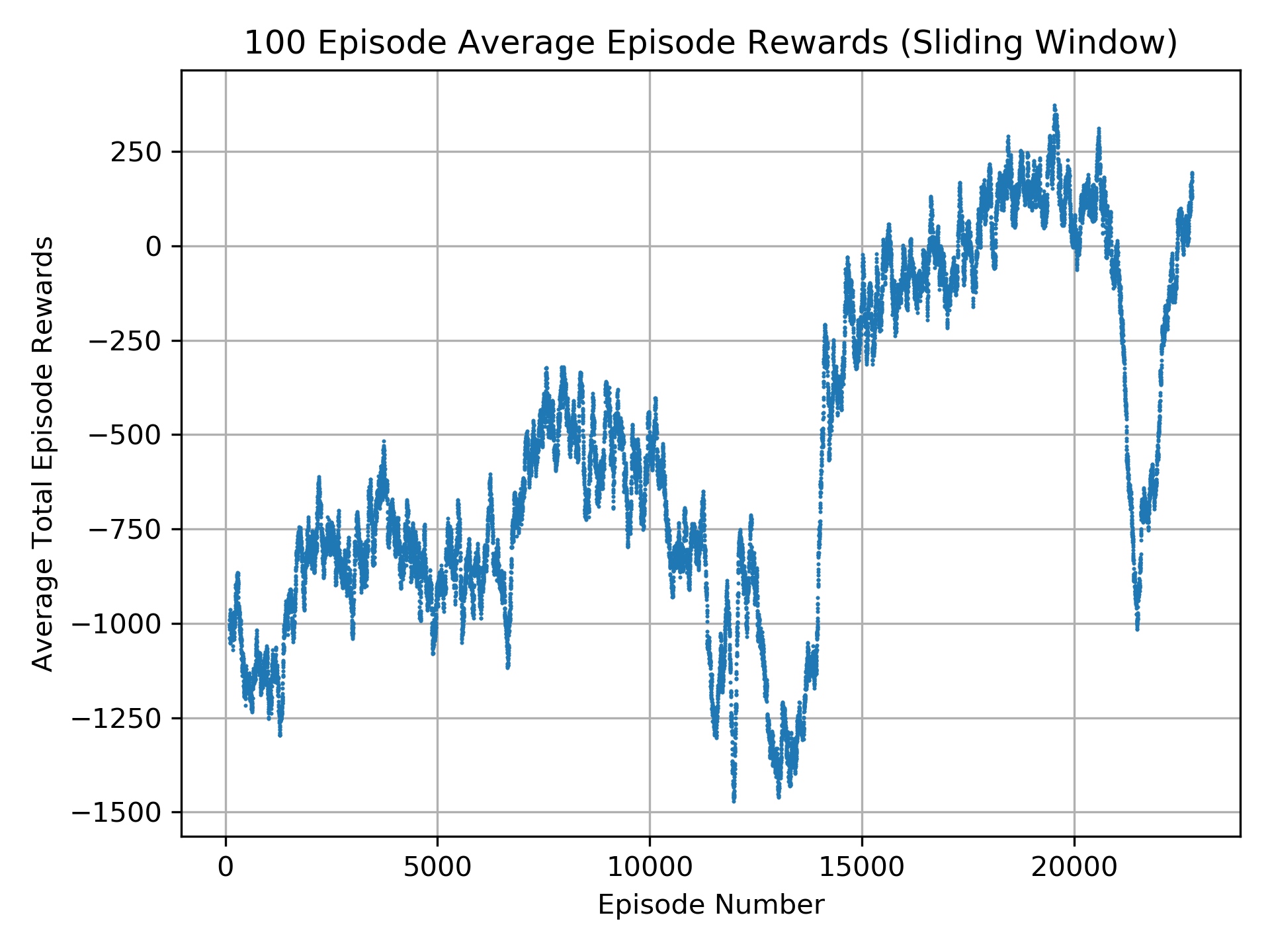} 
    \caption{Seed: 0} 
    \label{fig:200-reward-0} 
    % \vspace{1ex}
  \end{subfigure}
  \begin{subfigure}[b]{\linewidth}
    \centering
    \includegraphics[width=\linewidth]{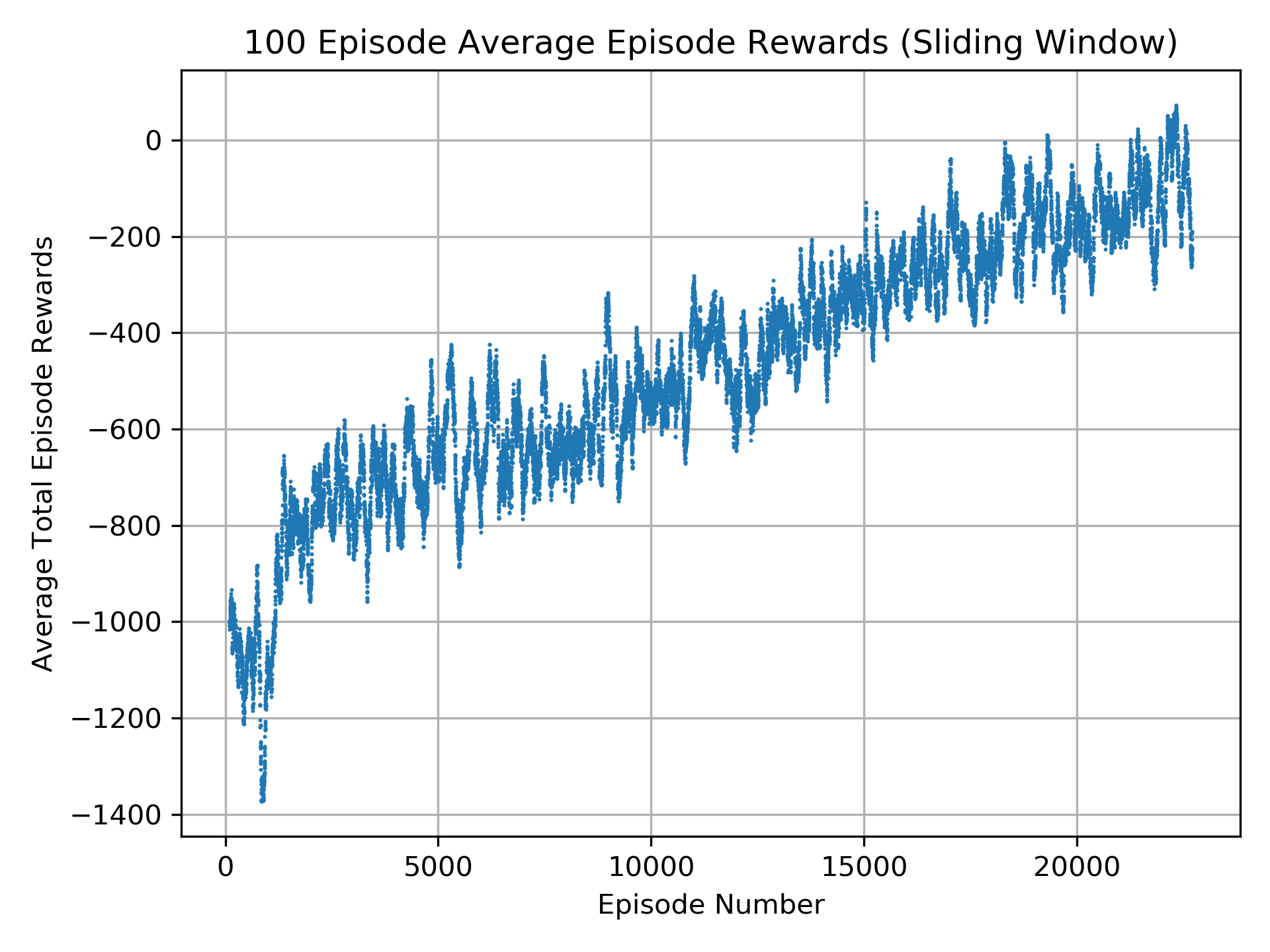} 
    \caption{Seed: 1} 
    \label{fig:200-reward-1} 
    % \vspace{1ex}
  \end{subfigure} 
  \caption{100 episode average episode rewards (sliding window), 200 bus system, no contingencies, different random seeds}
  \label{fig:200-reward} 
\end{figure}

Fig. \ref{fig:200-reward} presents average training rewards over time for two different random number generator seeds and the 200 bus system. Note that Fig. \ref{fig:200-reward-0} exhibits large reward fluctuations while Fig. \ref{fig:200-reward-1} shows steady improvements. This may be indicative of an oversized neural network, sub-optimal neural network architecture, and/or a need for hyper parameter tuning.

\begin{figure}[h!]
  \begin{subfigure}[b]{\linewidth}
    \centering
    \includegraphics[width=\linewidth]{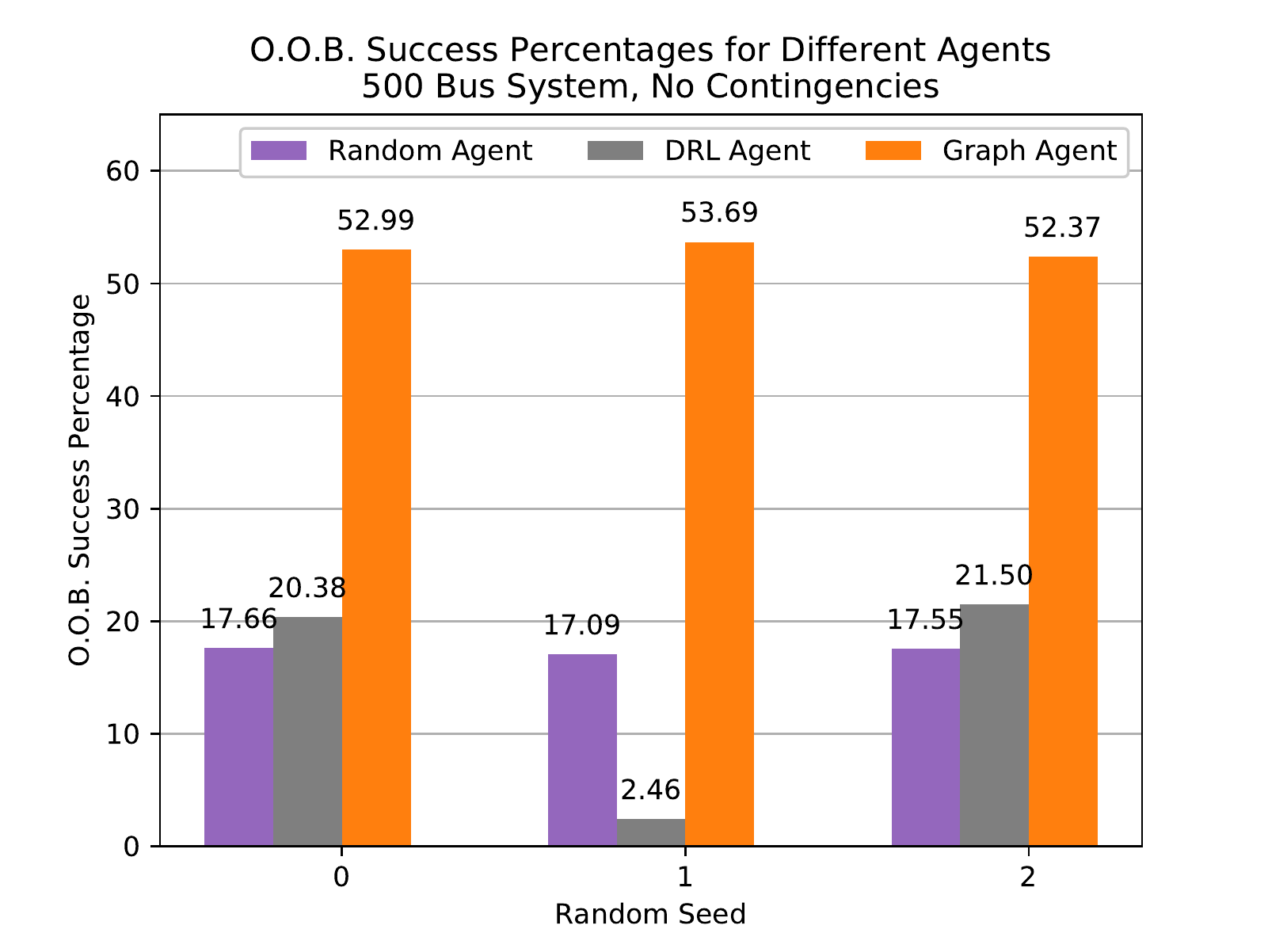} 
    \caption{No contingencies} 
    \label{fig:500-no-con} 
    % \vspace{1ex}
  \end{subfigure}
  \begin{subfigure}[b]{\linewidth}
    \centering
    \includegraphics[width=\linewidth]{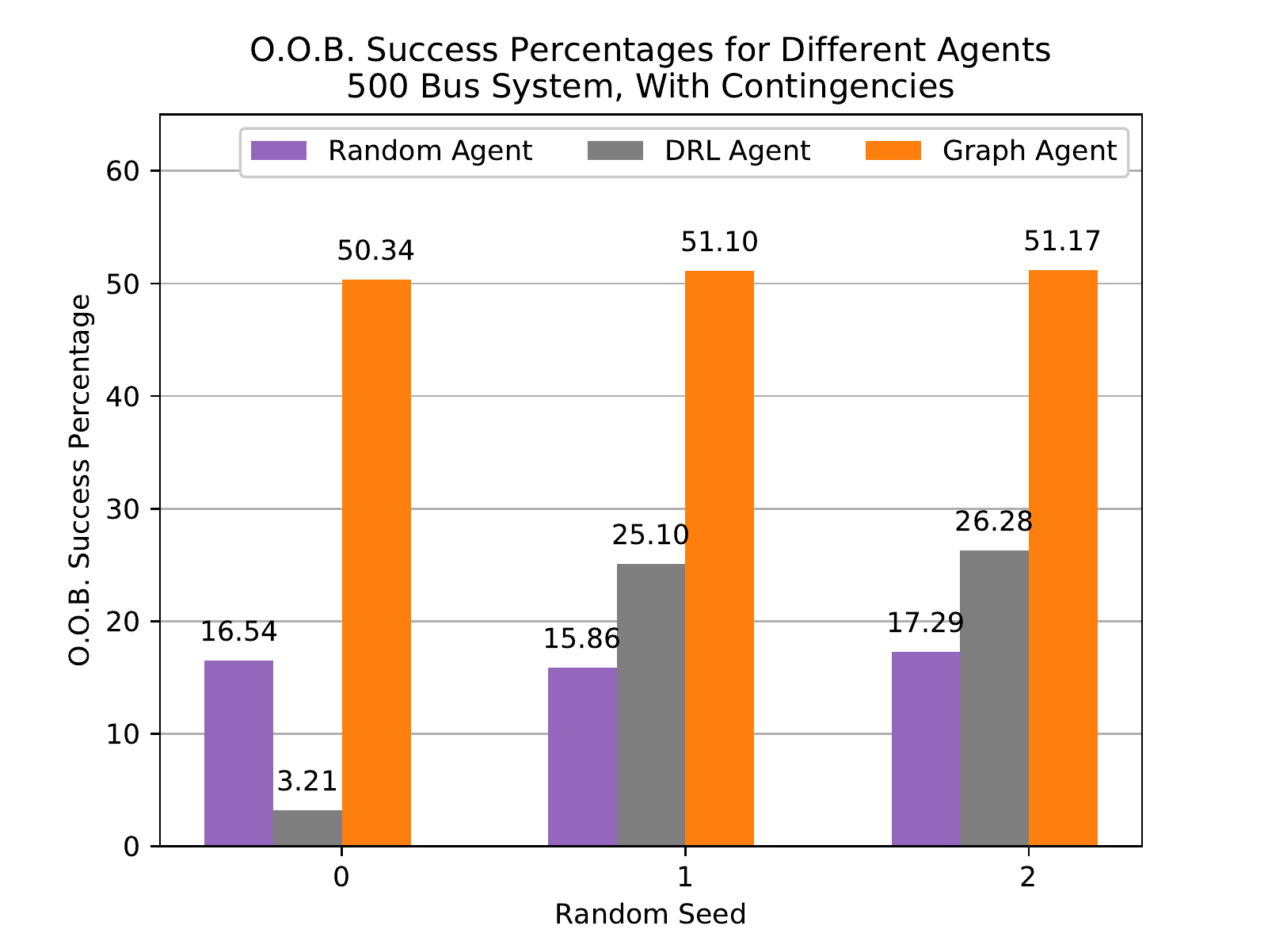} 
    \caption{With contingencies} 
    \label{fig:500-with-con} 
    % \vspace{1ex}
  \end{subfigure} 
  \caption{OOB success rates for random, DRL, and graph-based agent across different random seeds, 500 bus system, with and without contingencies}
  \label{fig:500-results} 
\end{figure}

Fig. \ref{fig:500-results} presents OOB success rates in the same fashion as Fig. \ref{fig:200-results}, but for the 500 bus system experiments. Interestingly, the random and graph-based agents exhibit significantly higher success rates than for the 200 bus system. This fact is likely related to structural differences between the two systems. Training instabilities can be seen in both Fig. \ref{fig:500-no-con} and Fig. \ref{fig:500-with-con} as each depicts a single instance of the DRL agent performing significantly worse than the corresponding random agent. Similar to the results for the 200 bus system, the presence of contingencies generally improved DRL agent performance, likely due to shortcomings in neural network architecture.

\section{Conclusions and Future Work} \label{sec:conclusion}

This paper provided an in-depth exploration of the application of deep reinforcement learning to the electric power transmission system voltage control problem. It was found that a novel deep-Q network algorithm modification wherein the agent is not allowed to take the same action multiple times in any given training or testing episode leads to significant DRL performance improvements. Additionally, it was shown that the min-max scaling of bus voltage observations could lead to performance improvements as opposed to simply using per unit voltages. DRL agents were trained to control 200 and 500 bus power systems in order to prove the scalability of DRL for voltage control. While no DRL agents were able to exceed the performance of the graph-based agents which were developed for comparison with DRL agents, there were cases with both the 14 and 200 bus systems where DRL agent performance approached graph-based agent performance. Training instabilities were observed for the larger test systems. The research presented in this paper shows clear potential for using DRL to solve the voltage control problem, but more work is needed to ensure DRL techniques can consistently outperform conventional techniques.

Opportunities for future work are numerous. An in-depth hyper parameter tuning study is likely necessary to get the most out of the advanced DQN algorithms used in this work. One intriguing prospect is the use of geographically arranged observations and convolutional neural networks (CNNs) instead of fully connected networks in the DQN algorithm. Alternatively, custom neural network structures that better exploit the local nature of power system voltage issues could improve performance. Finally, it would be valuable to compare DRL performance with state-of-the-art optimization-based voltage control schemes. 

% if have a single appendix:
%\appendix[Proof of the Zonklar Equations]
% or
%\appendix  % for no appendix heading
% do not use \section anymore after \appendix, only \section*
% is possibly needed

% use appendices with more than one appendix
% then use \section to start each appendix
% you must declare a \section before using any
% \subsection or using \label (\appendices by itself
% starts a section numbered zero.)
%

% Can use something like this to put references on a page
% by themselves when using endfloat and the captionsoff option.
\ifCLASSOPTIONcaptionsoff
  \newpage
\fi

% trigger a \newpage just before the given reference
% number - used to balance the columns on the last page
% adjust value as needed - may need to be readjusted if
% the document is modified later
%\IEEEtriggeratref{8}
% The "triggered" command can be changed if desired:
%\IEEEtriggercmd{\enlargethispage{-5in}}

% references section

% can use a bibliography generated by BibTeX as a .bbl file
% BibTeX documentation can be easily obtained at:
% http://mirror.ctan.org/biblio/bibtex/contrib/doc/
% The IEEEtran BibTeX style support page is at:
% http://www.michaelshell.org/tex/ieeetran/bibtex/
\bibliographystyle{IEEEtran}
% argument is your BibTeX string definitions and bibliography database(s)
\bibliography{citations}
\end{document}